\documentclass[11pt]{article}
\usepackage[preprint]{acl}
\usepackage{times}
\usepackage{latexsym}
\usepackage[T1]{fontenc}
\usepackage[utf8]{inputenc}
\usepackage{microtype}
\usepackage{graphicx}
\usepackage{amsmath,amssymb,amsfonts}
\usepackage{booktabs,multirow}
\usepackage{xcolor}
\usepackage{subcaption}
\usepackage{sidecap}
\usepackage{array}
\usepackage{tabularx}
\usepackage{makecell}
\usepackage{siunitx}
\usepackage{algorithm}
\usepackage{algorithmic}
\usepackage{url}
\usepackage{enumitem}

\newcommand{\ourmodel}{{\textsc{Weft}{}}}
\newcommand{\ourdataset}{{\textsc{MedFabric}}}
\newcolumntype{Y}{>{\raggedright\arraybackslash}X}
\renewcommand{\arraystretch}{0.95}

\title{\ourdataset{}: Gold Evidence Hides the Difficulty of Word-Level Medical Fabrication Detection}
\author{
\normalfont
Tung Sum Thomas Kwok$^{1}$ \quad
Qian Qian$^{2}$ \quad
Xiaofeng Lin$^{1}$ \quad
Dongxu Zhang$^{2}$ \quad
Jun Han$^{2}$ \\
Zhichao Yang$^{2}$ \quad
Davin Hill$^{2}$ \quad
Tamer Soliman$^{2}$ \quad
Sanjit Singh Batra$^{2}$ \quad
Robert Tillman$^{2}$ \quad
Guang Cheng$^{1}$ \\[4pt]
$^{1}$University of California, Los Angeles CA 90095, USA \quad
$^{2}$Optum AI, USA
}

\begin{document}
\maketitle

\begin{abstract}
Large language models fabricate in medicine, producing fluent statements that are factually wrong, so reliable fabrication detection is a prerequisite for clinical deployment. Reported progress on this task is inflated by two evaluation artifacts: an authorship-style shortcut, where human-written ground truths are paired with LLM-written hallucinations so detectors key on writing style rather than facts, and the provision of gold evidence at test time. A benchmark that tests factual reasoning must therefore remove the style shortcut, ground every fabrication in a real retrievable passage, and score detectors across the range of evidence quality faced in deployment. We build \ourdataset{} to these requirements, a benchmark of 646 word-level medical fabrications, each paired with a ground truth that shares its LLM authorship and near-identical surface form (median ROUGE-L 0.95). On \ourdataset{} the task is unsolved: expert clinicians reach only 53.3\% macro F1 and no detector family clears about 60\% without gold evidence. Our central finding is that detection is governed by evidence \emph{correctness} rather than fabrication subtlety, since a strong LLM scores 91\% with the gold passage but falls to 35\%, below its own no-evidence baseline, under a wrong one, a pattern that holds on two benchmarks at two model scales. The failure is actionable: a retrieval-confidence gate that abstains on low-confidence evidence raises macro F1 from 61\% to 74\%, and we release \ourdataset{}, all code, and every baseline.

\end{abstract}

\section{Introduction}
\label{sec:intro}
\begin{figure*}[t]
    \centering
    \includegraphics[width=\textwidth]{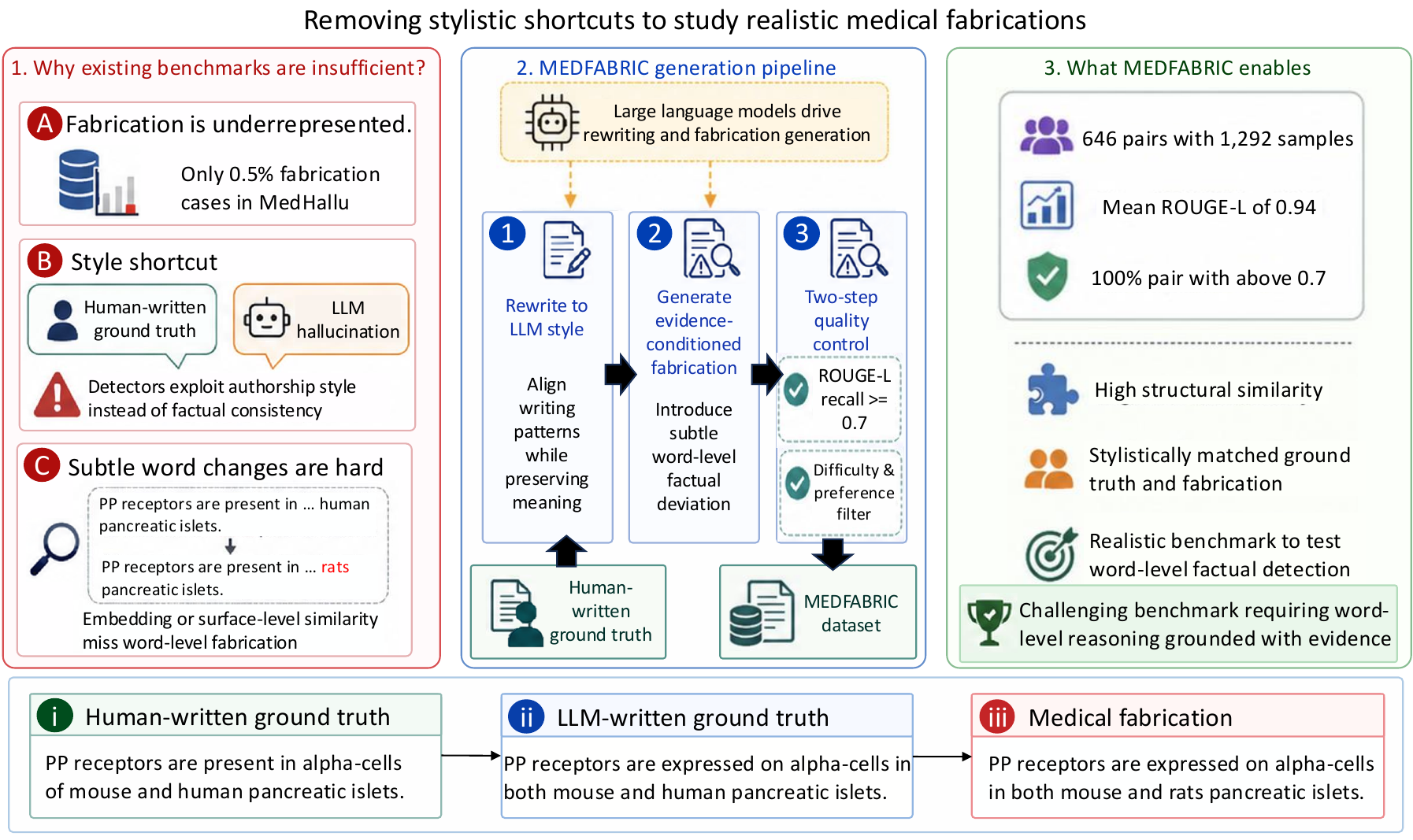}
    \caption{Motivation and overview of \ourdataset{}. (1)~Why existing benchmarks are insufficient: fabrication is underrepresented, detectors exploit authorship style rather than factual consistency, and subtle word-level changes evade embedding- or surface-level similarity. (2)~The \ourdataset{} generation pipeline: rewrite to LLM style, generate evidence-conditioned fabrications, and apply two-step quality control. (3)~What \ourdataset{} enables: a stylistically matched, high-similarity benchmark for word-level factual detection grounded in evidence. The bottom row traces one example from human-written ground truth, to LLM-rewritten ground truth, to a subtle medical fabrication.}
    \label{fig:generation-overview}
\end{figure*}
Large Language Models (LLMs) have demonstrated strong semantic understanding, in-context learning, and logical reasoning~\cite{kamalloo-etal-2023-evaluating}, but they routinely hallucinate in domains that demand expert knowledge~\cite{10444954}. The most dangerous form is \emph{fabrication}: the generation of factually incorrect but fluent statements~\cite{10.1145/3703155}, which carries acute risk in medicine~\cite{Kim2025.02.28.25323115}. Reliably \emph{detecting} such fabrications is therefore a prerequisite for trustworthy medical LLM deployment.

We argue that current evaluations substantially \emph{overstate} how well medical fabrications can be detected, owing to two artifacts. First, an \textbf{authorship-style shortcut}: existing medical hallucination datasets (e.g., Med-HALT, MedHallu~\cite{pal-etal-2023-med,pandit2025medhallucomprehensivebenchmarkdetecting}) pair \emph{human}-written ground truths with \emph{LLM}-written hallucinations, so a detector can separate the two by writing style rather than factual content~\cite{zhang2024knowhaluhallucinationdetectionmultiform}. Second, \textbf{gold-evidence provision}: detectors are handed the exact supporting passage at test time, reducing fabrication detection to an entailment check that does not reflect deployment, where evidence must be retrieved and is frequently wrong. Table~\ref{tab:benchmark-comparison} contrasts \ourdataset{} with prior medical benchmarks on the dimensions these artifacts turn on.

\smallskip\noindent\textbf{Research questions.} Removing these two artifacts leaves open what detection performance actually is, which we frame as three questions:
\begin{itemize}[leftmargin=*,itemsep=1pt,topsep=2pt]
\item In the realistic setting where no gold evidence is supplied, how well can current detectors, and expert clinicians, identify word-level medical fabrications? (Section~\ref{sub:evidence}, Section~\ref{sub:human})
\item What governs detection difficulty: the subtlety of the fabricated edit, or the \emph{correctness} of the evidence supplied to the detector? (Section~\ref{sub:evidence})
\item Given this failure mode, can detection reliability be recovered without retraining the detector? (Section~\ref{sub:mitigation})
\end{itemize}

\smallskip\noindent\textbf{Design considerations.} Answering these questions requires a benchmark that forces detectors to reason about \emph{facts} rather than artifacts, which imposes three requirements. First, it must eliminate the authorship-style shortcut, so that a fabrication is not betrayed by its writing style. Second, it must ground each fabrication in a real, retrievable evidence passage rather than a synthetic template. Third, it must score detectors across the range of evidence quality faced in deployment, not only with a gold passage. \ourdataset{} is designed to these requirements, realised as Goals G1 to G4 in Section~\ref{sec:benchmark}.

To remove the first artifact, we build \ourdataset{}, a benchmark of \textbf{word-level} medical fabrications whose ground truths and fabrications \emph{share} LLM authorship and near-identical surface form (median ROUGE-L 0.95, Figure~\ref{fig:generation-overview}). A three-stage pipeline rewrites human answers into LLM style, generates evidence-conditioned fabrications that differ by a single edit (an entity, number, negation, or predicate, for example flipping mortality risk in patients \emph{with} to \emph{without} head injuries, the running example we revisit in Section~\ref{sub:cases}), and filters for structural and semantic similarity, producing pairs that differ by a genuine factual change but not by style (Figure~\ref{fig:similar-embedding}).

On \ourdataset{}, the task is hard for \emph{every} method we test. Expert clinicians given the retrieved evidence reach only 53.3\% macro F1, near chance, and a fine-tuned DeBERTa collapses to 33.3\% once the style shortcut is removed. No detector we evaluate exceeds about 60\% macro F1 without gold evidence. This holds for naive LLM-as-judge, a structured evidence-grounded pipeline (\ourmodel{}, Section~\ref{sub:ourmodel}), an off-the-shelf NLI verifier, and medical fine-tuned and reasoning models.

Our central finding addresses the second artifact and holds across two benchmarks and two model scales: \textbf{detection quality is governed by evidence \emph{correctness}, not fabrication subtlety}. A strong LLM handed the gold passage reaches 91\%, but that figure is an artifact of correct evidence: under \emph{wrong} (mis-retrieved) evidence, detection falls \emph{below chance} (to about 35\%), worse than using no evidence at all, because the detector trusts the passage and is misled by it. Benchmarks that supply gold evidence therefore report an upper bound that collapses under realistic retrieval.

Finally, the failure is \textbf{actionable}. Because detection fails exactly when evidence is unreliable, a one-line \textbf{retrieval-confidence gate}, the question-evidence similarity, that abstains on low-confidence cases restores near-gold reliability (61\% to 74\% macro F1) by deferring wrong-evidence cases to human review.\footnote{\ourdataset{}, all code, and every baseline are released through an anonymized repository: \url{https://anonymous.4open.science/r/medfabric-74D2/README.md}}

  \subsubsection{Contributions}
  \begin{itemize}[leftmargin=*,itemsep=1pt,topsep=2pt]
  \item \textbf{Benchmark.} We release \ourdataset{}, a 1,292-sample benchmark of word-level medical fabrications (median ROUGE-L 0.95) whose paired answers share LLM authorship, removing the authorship-style shortcut so that detection must rest on facts rather than style (Section~\ref{sec:benchmark}).
  \item \textbf{Evaluation.} We show the task is unsolved without gold evidence, where no detector family clears about 60\% macro F1 and expert clinicians sit at 53.3\%, and that detection is governed by evidence \emph{correctness} rather than fabrication subtlety, since gold evidence yields 91\% while wrong evidence drives detection below chance to about 35\% on two benchmarks and two model scales (Section~\ref{sub:evidence},~Section~\ref{sub:human}).
  \item \textbf{Actionable mitigation.} We turn this failure into a deployment recipe: a model-free retrieval-confidence gate that abstains on unreliable evidence recovers near-gold reliability (61\% to 74\%), and we release the paired gold/none/distractor conditions so that detectors can be trained against the wrong-evidence collapse, which we leave to future work (Section~\ref{sub:mitigation}, Appendix~\ref{app:opportunities}).
\end{itemize}

\section{Related Work}
\label{sec:related}
\noindent\textbf{Medical hallucination benchmarks.} LLM hallucination, the generation of factually incorrect or nonsensical content~\cite{10.1145/3703155}, has been probed by a succession of datasets, from the true/false statements of \citet{azaria-mitchell-2023-internal} to conversational benchmarks such as HaluEval~\cite{li-etal-2023-halueval} and the multi-type taxonomies built on them~\cite{zhang2024knowhaluhallucinationdetectionmultiform}. Three medical benchmarks are the closest points of comparison. Med-HALT~\cite{pal-etal-2023-med} and MedHallu~\cite{pandit2025medhallucomprehensivebenchmarkdetecting} both pair \emph{human}-written ground truths with \emph{LLM}-written hallucinations, leaving the authorship-style shortcut in place, and fabrication is scarce in both (0.5\% of the MedHallu corpus) despite being the failure mode that matters most in high-stakes domains~\cite{li2023chatdoctor}. MEDEC~\cite{medec2024} annotates word-level clinical errors but supplies no per-instance evidence passage, placing it outside the evidence-grounded setting we study. \ourdataset{} closes both gaps at once (Table~\ref{tab:benchmark-comparison}).

\begin{table}[t]
\centering
\footnotesize
\setlength{\tabcolsep}{3.5pt}
\resizebox{\columnwidth}{!}{%
\begin{tabular}{lcccccc}
\toprule
Benchmark & Fab. & Word & Style-free & Evid. & Human & Sweep \\
\midrule
Med-HALT & $\times$ & $\times$ & $\times$ & $\times$ & $\times$ & $\times$ \\
MedHallu & $\times$ & $\times$ & $\times$ & \checkmark & (\checkmark) & $\times$ \\
MEDEC & (\checkmark) & \checkmark & $\times$ & $\times$ & \checkmark & $\times$ \\
\ourdataset{} (ours) & \checkmark & \checkmark & \checkmark & \checkmark & \checkmark & \checkmark \\
\bottomrule
\end{tabular}}
\caption{\ourdataset{} versus prior medical benchmarks (Med-HALT~\cite{pal-etal-2023-med}, MedHallu~\cite{pandit2025medhallucomprehensivebenchmarkdetecting}, MEDEC~\cite{medec2024}) on the dimensions that decide whether detection tests \emph{factual} rather than stylistic or gold-evidence-aided ability. \emph{Fab.}: fabrication-focused. \emph{Word}: word-level edits. \emph{Style-free}: ground truth and fabrication share LLM authorship, with no authorship-style shortcut. \emph{Evid.}: per-instance evidence passage. \emph{Human}: expert human validation. \emph{Sweep}: evaluation across gold/retrieved/none/distractor evidence. (\checkmark)~= partial.}
\label{tab:benchmark-comparison}
\end{table}

\noindent\textbf{Fabrication detection.} Detectors fall into three families~\cite{huang2023factualinconsistencyproblemabstractive}: \emph{uncertainty-based} methods reading token probabilities, verbalized confidence, or sampling consistency~\cite{kuhn2023semantic,lin-etal-2022-truthfulqa,manakul-etal-2023-selfcheckgpt}, \emph{supervised} probes trained on labelled hallucinations under the assumption that truthfulness is encoded in internal states~\cite{azaria-mitchell-2023-internal}, and \emph{LLM-agentic} frameworks that use one or more LLMs as evaluators~\cite{huo2023retrievingsupportingevidencellms}. All three judge an answer largely in isolation, which is why the complementary line of \emph{retrieval-augmented verification}~\cite{shuster-etal-2021-retrieval-augmentation,jin-etal-2019-pubmedqa,10.1609/aaai.v39i22.34559} is the one our results bear on most directly: these methods treat retrieval as an input-augmentation step and implicitly assume the retrieved passage is correct, which is precisely the assumption we show detection depends on (Section~\ref{sub:evidence}). We evaluate representatives of every family in Section~\ref{sec:results}. A fuller treatment of each family, of how our \ourmodel{} baseline relates to prior evidence-grounded detectors, and of the human-in-the-loop validation literature is given in Appendix~\ref{app:relatedwork}.

\section{The \ourdataset{} Benchmark}
\label{sec:benchmark}
Existing medical hallucination datasets provide limited and low-quality coverage of fabrication cases, with only 0.5\% of the MedHallu corpus involving such instances~\cite{pandit2025medhallucomprehensivebenchmarkdetecting}. Fabrications represent a distinct data distribution, reflected by poor performance from detectors trained mainly on non-fabrication hallucinations when evaluating fabrication-specific samples (Appendix~Table~\ref{tab:fab-vs-non-fab}). This shows the need for a fabrication-focused benchmark, which we introduce as \ourdataset{} (Section~\ref{sub:ourdataset}). However, on this fabrication-focused benchmark most existing detectors degrade sharply, which we attribute to embedding models' inability to distinguish subtle word-level fabrications (Section~\ref{sub:ourmodel}). 
\subsection{Word-Level Fabrication Generation} \label{sub:ourdataset}
\noindent\textbf{Design scope.} \ourdataset{} is built to four goals. \textbf{(G1) Realism:} every fabrication is a genuine word-level LLM failure mode, an altered entity, number, negation, or predicate, grounded in real clinical QA rather than a synthetic template. \textbf{(G2) No stylistic shortcut:} ground truth and fabrication share LLM authorship and near-identical surface form, so no detector can win on writing style. \textbf{(G3) Deployment-faithful evaluation:} the benchmark is scored across evidence conditions (gold / retrieved / none / distractor) and fabrication types rather than as a single leaderboard number. \textbf{(G4) Quality and scalability:} an automated rewrite, fabricate, and filter pipeline with ROUGE-L/SPPO quality control and expert validation, cheaply extensible to new medical sources. The task is deliberately narrow: identify a single word-level factual error in an otherwise-correct answer.

\noindent\textbf{Curation.} \ourdataset{} is derived from MedHallu~\cite{pandit2025medhallucomprehensivebenchmarkdetecting}, a multi-domain medical QA benchmark spanning pharmacology, clinical oncology, rheumatology, and general internal medicine, with all three pipeline stages executed by GPT-4o~\cite{openai2024gpt4ocard}. Detection is formulated as binary classification: given a question, supporting evidence, and a candidate answer, predict ground truth or fabrication (Appendix~\ref{app:construction}).

\noindent\textbf{Existing bottleneck.} Existing generation pipelines substantially alter the linguistic form of the ground truth, so detectors reach inflated accuracy by telling writing styles apart rather than by checking facts: an online writing detector scores a MedHallu hallucination 72\% AI-written against 3\% for its human-written ground truth (Table~\ref{tab:medhallu-gptzero}). We therefore replace human-authored ground truths with LLM-rewritten counterparts, condition fabrications on those rewrites, and constrain every alteration to the word level, producing pairs with a median ROUGE-L of 0.95 that still induce misclassification in state-of-the-art models~\cite{katz2023gpt4}.

\noindent\textbf{Rewriting process.} We prompt the LLM to rewrite each human-written ground truth into LLM style without altering its semantic content (Figure~\ref{fig:generation-overview}), which aligns the stylistic distribution of the two classes and is what removes the authorship-style shortcut (Appendix~\ref{app:construction}, formal operator in Appendix~\ref{def:rewrite}).

\noindent\textbf{Fabrication process.} Conditioned on the supporting evidence and the rewritten ground truth, the LLM then introduces a single word-level factual error, with the evidence chunks and the correct answer supplied in the prompt so that grounding does not rely on the model's parametric knowledge (Appendix~\ref{def:fabricate}).

\noindent\textbf{Validation process.} Every candidate passes a two-step automatic filter (Algorithm~\ref{alg:fabrication_qc}): fabrications whose ROUGE-L recall~\cite{rouge} against the ground truth falls below 0.7~\cite{Faray_de_Paiva2025.02.06.25321749} are discarded as structurally divergent, and a preference-based difficulty filter~\cite{wu2024selfplaypreferenceoptimizationlanguage} retains only fabrications that an LLM judge wrongly prefers over the ground truth, keeping the benchmark non-trivial. Because automatic filters cannot certify a \emph{genuine} factual error, three biomedically-trained annotators also reviewed a stratified 120-pair sample (18.6\% of the benchmark) blind to the labels. After adjudicating the 73 disputed pairs against the full source document, 85.8\% of sampled fabrications were confirmed as genuine factual errors, the residual 14\% being \emph{unsupported additions} that mark the benchmark's construct boundary (full protocol in Appendix~\ref{app:human}).

\begin{table}[t]
\centering
\renewcommand{\arraystretch}{1.1}
\setlength{\tabcolsep}{4pt}
\resizebox{\linewidth}{!}{%
\begin{tabular}{l rr rrr}
\toprule
\textbf{Dataset} & \textbf{\#Pairs} & \textbf{Train / Test} & \textbf{Mean ROUGE-L} & \textbf{\% $\geq$0.7} & \textbf{Source} \\
\midrule
MedHallu~\cite{pandit2025medhallucomprehensivebenchmarkdetecting} & 9,670 & 80 / 20 & 0.12 & 0.4\% & Human-authored \\
\ourdataset{} (v1)  & 258  & 70 / 30 & 0.23 & 1.2\%  & LLM-rewritten \\
\ourdataset{} (v2.2) & 646  & 60 / 40 & \textbf{0.87} & \textbf{85\%} & LLM-rewritten \\
\bottomrule
\end{tabular}}
\caption{\ourdataset{} statistics compared with MedHallu. Each pair consists of one ground-truth and one fabricated answer for the same question (v2.2: 646 pairs = 1,292 samples total). Version 2.2 applies a ROUGE-L recall quality-control target of at least 0.7, satisfied by 85\% of released pairs (median ROUGE-L 0.95, and the mean 0.87 is lowered by a minority of longer added-clause fabrications), producing structurally near-identical fabrication pairs that challenge all surface-level detectors.}
\label{tab:dataset-stats}
\end{table}

\subsection{Benchmark Characteristics}
\label{sub:characteristics}
\begin{figure}[t]
\centering
\includegraphics[width=\columnwidth]{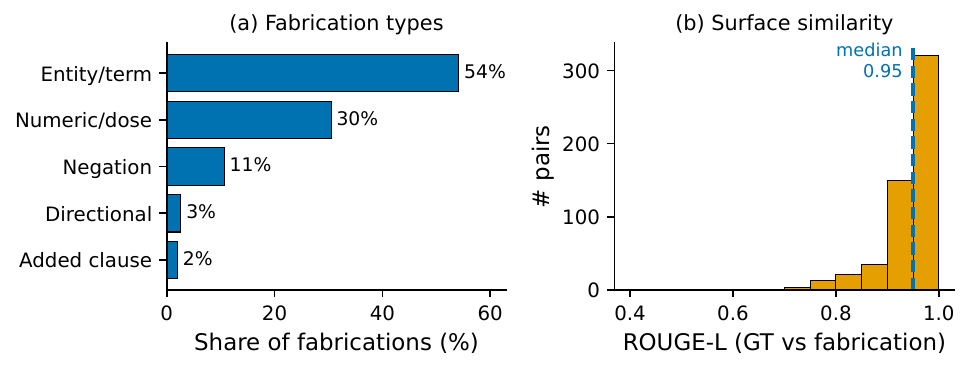}
\caption{\ourdataset{} characteristics. (a)~Fabrication-type distribution: entity/term and numeric edits dominate, with a negation/directional tail. (b)~Ground-truth vs.\ fabrication ROUGE-L: most pairs are near-identical (median 0.95), so each fabrication is a genuine word-level change rather than a stylistic rewrite. Per-type detectability is analysed in Section~\ref{sub:taxonomy}.}
\label{fig:characteristics}
\end{figure}

\ourdataset{} v2.2 contains 646 fabrication and ground-truth pairs (1,292 answers) with a 60/40 train/test split and no separate development split, so 387 pairs (774 answers) are available for exemplar selection and 259 pairs (518 answers) are held out for evaluation, drawn from pharmacology, clinical oncology, rheumatology, and general internal medicine (scale in Table~\ref{tab:dataset-stats}, feature comparison in Table~\ref{tab:benchmark-comparison}). Two properties define its difficulty (Figure~\ref{fig:characteristics}). First, fabrications are dominated by entity/term substitutions (54\%) and numeric/dose changes (30\%), with subtler negation and directional flips forming a long tail. Second, every fabrication is a near-minimal edit of its ground truth: the median GT-to-fabrication ROUGE-L is 0.95 and 85\% of pairs clear the 0.7 quality-control threshold, so surface- and embedding-level cues carry little signal (Section~\ref{sub:ourmodel}) and detection must rest on the single altered fact.

\subsection{Detectors and the \ourmodel{} baseline} \label{sub:ourmodel}
\begin{figure}[t]
  \centering
  \includegraphics[width=\columnwidth]{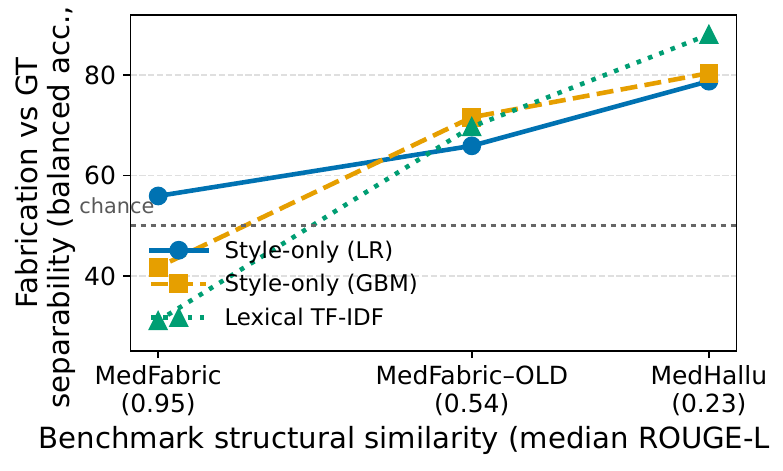}
  \caption{Detectors that key on style or lexis lose their signal as fabrications approach their ground truth. A classifier using surface features alone separates fabrication from ground truth at 80 to 88\% on the low-similarity MedHallu pairing, but collapses toward or below chance (31 to 56\%) on \ourdataset{}, where both answers share LLM authorship and median ROUGE-L is 0.95 (5-fold CV, Appendix~\ref{app:shortcut}).}
  \label{fig:existing-performance}
\end{figure}
Existing fabrication detectors degrade sharply as the structural similarity between ground truth and fabrication rises (Figure~\ref{fig:existing-performance}): they largely key on stylistic and structural cues rather than factual consistency, so on a high-similarity benchmark like \ourdataset{} they lose their signal. This overfitting to surface form is clearest for embedding-only detectors: off-the-shelf embedding models are near-random on \ourdataset{} (41 to 48\% macro F1) because the two answers have almost identical embeddings (Figure~\ref{fig:similar-embedding}), while zero-shot LLM judges do better but stay sensitive to sampling and answer order (Section~\ref{sub:committee}). As one baseline we add \ourmodel{}, a zero-shot structured detector that decomposes each answer into entity-description rows, refills every content word from the retrieved evidence, and scores the reconstruction with an embedding-distance gate followed by an LLM verdict, so that word-level divergence from the evidence is made explicit rather than left latent (full pipeline, verbatim-copy audit, and embedding-gate analysis in Appendix~\ref{app:weft}). On \ourdataset{} this machinery only matches a one-line few-shot prompt (Table~\ref{tab:leaderboard}), so we report \ourmodel{} as a baseline rather than a contribution.

\section{Results and Analysis}
\label{sec:results}
Our experiments address the three research questions posed in Section~\ref{sec:intro} through four
findings. Firstly, \textbf{we find that detection is decided by the evidence supplied rather than by
the detector}, since no method clears about 60 macro F1 without a passage while correct evidence
lifts detection to 68 to 91 (Section~\ref{sub:evidence}). Secondly, \textbf{we show that a single-word
negation flip is the hardest fabrication to catch}, inverting the order of how common each
fabrication type is (Section~\ref{sub:taxonomy}). Thirdly, \textbf{we recover most of the lost reliability
without retraining}, through a retrieval-confidence gate that abstains when the evidence is
unreliable (Section~\ref{sub:mitigation}). Finally, \textbf{we establish that the difficulty belongs to the
fabrications rather than to the judge that selected them}, holding across an order-invariant
committee at two model scales (Section~\ref{sub:committee}). We also measure an expert human ceiling that
is itself near chance, bounding what any snippet-grounded detector can achieve (Section~\ref{sub:human}).

\subsection{Experimental setup}
\label{sub:setup}
\noindent\textbf{Task and metric.} Every detector receives a question and a candidate answer (plus a supporting passage where the method consumes evidence) and outputs a binary \texttt{gt}/\texttt{fabricated} label, scored by macro F1 on the held-out test split of \ourdataset{} v2.2 (646 pairs, 1,292 answers). The benchmark is class-balanced, so macro F1 weights both classes equally and a degenerate single-class predictor scores only 33.3. All margins carry paired-bootstrap 95\% confidence intervals (Appendix~\ref{app:bootstrap}).
\textbf{Baselines.} We cover families and scales: a frontier LLM (GPT-4o), open LLMs at two scales (Qwen3-4B, Qwen3-30B), a reasoning model (DeepSeek-R1-Distill-Qwen-7B), a medical fine-tuned model (BioMistral-7B), supervised encoders (a fine-tuned DeBERTa and an attention probe), structured and graph detectors (\ourmodel{}, GCA, TSV), an off-the-shelf NLI verifier (DeBERTa-MNLI), and expert humans.
\textbf{Protocol.} LLMs are run 5-shot with exemplars drawn only from train, and we sweep the evidence handed to the judge across gold / retrieved / none / distractor (Section~\ref{sub:evidence}). Three biomedically-trained annotators label a stratified 120-pair sample blind to the labels (Section~\ref{sub:human}). Model versions, decoding, prompts, and the exact scoring rule are in Appendix~\ref{app:baselines}.

\subsection{Detection depends on provided evidence.}
\label{sub:evidence}
\begin{figure}[t]
\centering
\includegraphics[width=\columnwidth]{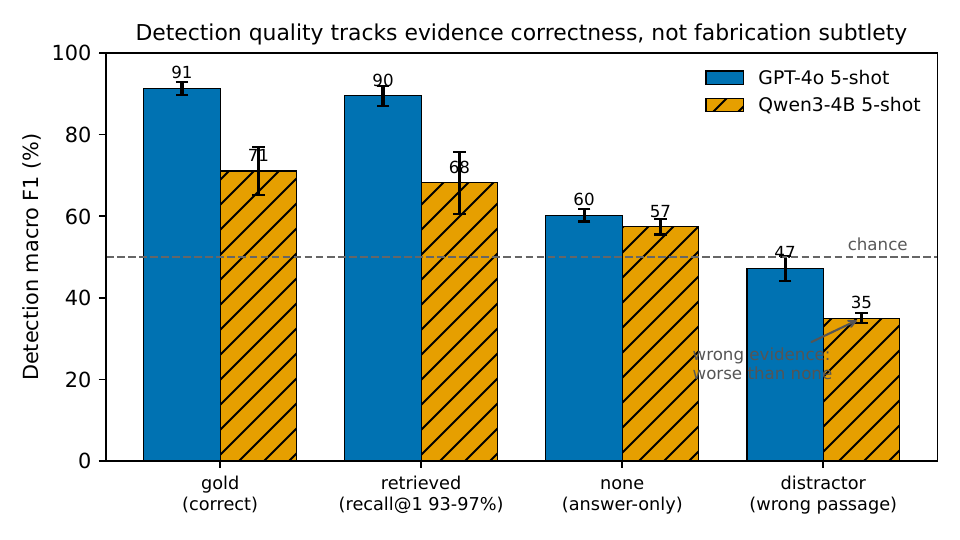}
\caption{Detection macro F1 vs.\ evidence quality (5-shot, 3 seeds, error bars = std). Correct
evidence (gold/retrieved) carries the task, while \emph{wrong} evidence pushes both models below the
no-evidence baseline, below chance for Qwen3-4B. Full per-condition table in Appendix~\ref{app:evsweep}.}
\label{fig:evsweep}
\end{figure}
Table~\ref{tab:leaderboard} shows each detector family in the deployable setting, where the judge has no access of the gold evidence passage. We observe that no method, not even expert clinicians (Section~\ref{sub:human}) reaches a macro F1 above 60. Specifically, this failure is indifferent to detector design as the supervised encoders fail once the authorship-style shortcut is removed, which matched pairing DeBERTa degenerating to a single-class predictor (33.3) and other models having only about 40\% accuracy. Stronger detectors are few-shot LLMs with our structured \ourmodel{} pipeline matching a one-line few-shot prompt. Importantly, neither remedy a reader might reach for helps, as the medical fine-tuned BioMistral-7B shows to be the weakest LLM, which corroborates the observation of \citet{pandit2025medhallucomprehensivebenchmarkdetecting} that domain tuning does not support hallucination detection. It is not the model's reasoning ability that serves as the bottleneck of medical fabrication detection, but the sufficiency of internal medical knowledge to judge the claim on its own. 

\begin{table}[t]
\centering
\small
\resizebox{\columnwidth}{!}{%
\begin{tabular}{lcl}
\toprule
Detector & Macro F1 & Evidence to judge \\
\midrule
GPT-4o 5-shot & \textbf{60.2}\,\tiny{$\pm$1.5} & none (answer-only) \\
Qwen3-4B 5-shot & \underline{57.4}\,\tiny{$\pm$1.9} & none (answer-only) \\
\ourmodel{} (Qwen3-4B) & 56.7 & evidence at fill step$^\dagger$ \\
NLI (DeBERTa-MNLI, evidence) & 56.6 & gold evidence \\
Expert humans (majority) & 53.3 & retrieved snippet \\
R1-Distill-Qwen-7B (reasoning) & 51.6\,\tiny{$\pm$2.3} & none (answer-only) \\
LLM-as-judge (naive) & 47.0 & none \\
GCA & 47.2 & knowledge graph \\
TSV & 44.1 & none \\
Attention probe (supervised) & 36.8 & hidden states \\
BioMistral-7B (medical fine-tuned) & 34.3\,\tiny{$\pm$1.8} & none (answer-only) \\
DeBERTa (matched LLM-vs-LLM) & 33.3 & pair input \\
\bottomrule
\end{tabular}}
\caption{Detection on \ourdataset{} v2.2 in the realistic (no-gold-evidence-to-judge) regime.
No detector clears about 60. The structured \ourmodel{} pipeline does not beat a plain few-shot
prompt. Best in \textbf{bold}, \underline{second-best} underlined. $^\dagger$\ourmodel{} consumes evidence only at its reconstruction step, not at the verdict.}
\label{tab:leaderboard}
\end{table}

In the contrary, by providing the knowledge, Figure~\ref{fig:evsweep} shows the detector performance improves significantly. We provide evidence of various quality to both small open source Qwen3-4B~\cite{qwen3} and proprietary GPT-4o~\cite{katz2023gpt4}, including  \textbf{gold}, \textbf{retrieved}, \textbf{none}, and a deliberately wrong \textbf{distractor} passage. We observe that given a gold or successfully-retrieved evidence, detection accuracy is at 68 - 91\%, while taking it out returns the performance to 57 - 60\%, resonating results in Table~\ref{tab:leaderboard}. The relationship between evidence quality and detector performance is further supported by the degeneration when the distractor passage is inputed. Appendix~\ref{app:drivers} further shows that neither editing subtlety nor its overlap with the evidence can predict detection, and the same degradation appears at both scales on \textbf{MedHallu}~\cite{pandit2025medhallucomprehensivebenchmarkdetecting}, which is an independently-constructed benchmark with the authorship-style shortcut. This suggests that the dependence is a property of the detection task instead of one dataset's construction. We include further information regarding the analysis in Appendix~\ref{app:evsweep}. \textbf{Finding~1:} LLMs lack the internal medical knowledge this task demands and relies on the evidence they are given, so detection rises and falls with the quality of that
evidence.

\subsection{Hardest fabrication types}
\label{sub:taxonomy}
\ourdataset{}'s fabrication types' distribution and difficulty are uneven and negatively correlated. Entity or term substitutions dominate at 54\% of the benchmark and numeric or dosage changes account for a further 31\%, with negation flips (11\%) and directional or comparator swaps (3\%) forming a long tail (Figure~\ref{fig:characteristics}a). The catch rate of each type from a representative evidence-grounded detector invert that order however (Appendix~\ref{app:taxonomy}), as single-token \emph{negation flips} are the hardest to catch at 58.8\%, entity swaps and added unsupported clauses sit in the mid-60s, and numeric and directional edits are the easiest at 76.5\% and 82.4\%. The ordering follows from what each edit leaves for a detector to check. A numeric or directional edit substitutes a token that the passage either confirms or contradicts almost mechanically, since a dose appears in the evidence or it does not and a reversed comparator runs against the direction the evidence states. A negation flip changes one function word, leaves every content word and the sentence topic intact, and alters only the polarity of the claim, which a detector recovers solely by reading the evidence for what it asserts but not which entities it mentions. \textbf{Finding~2:} Single-word negation flips are the hardest fabrications to catch.

\subsection{Retrieval confidence improves prediciton reliability}
\label{sub:mitigation}
\begin{figure}[t]
\centering
\includegraphics[width=0.96\columnwidth]{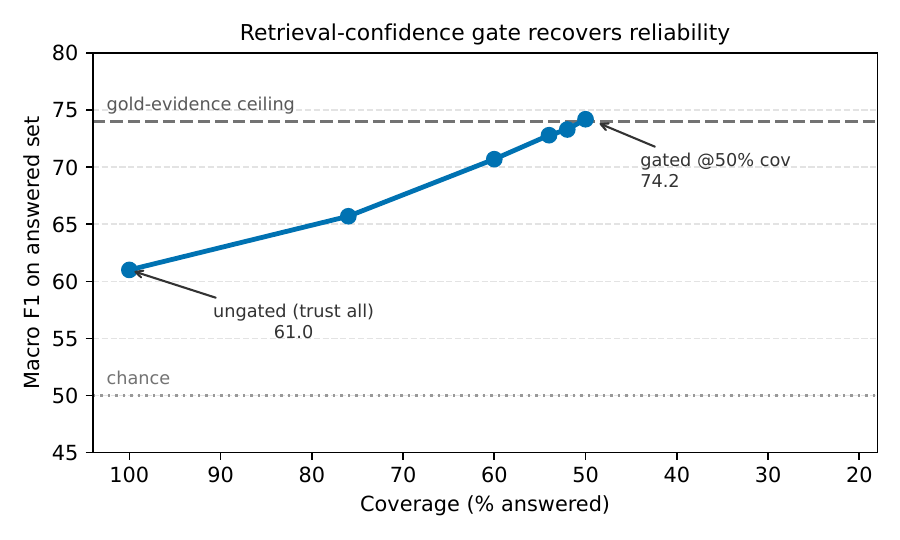}
\caption{Risk-coverage under mixed (gold/distractor) evidence. Abstaining on low retrieval-confidence
cases raises macro F1 on the answered set from the ungated 61.0 toward the gold-evidence ceiling
(about 74) as coverage drops, the wrong-evidence cases, which fail below chance, are deferred to
human review. Full risk-coverage table in Appendix~\ref{app:mitigation}.}
\label{fig:mitigation}
\end{figure}
With the result depending on evidence, we propose an actionable deployment. If detection fails consistently when the provided evidence is wrong, a detector should \emph{abstain} on low-confidence evidence and route those cases to human review. We test the simplest possible gate, retrieval confidence measured as the TF-IDF cosine similarity between the question and the provided passage, requiring no extra model. On a pool that mixes correct (gold) and wrong (distractor) evidence in equal parts, modelling a retriever that is right half the time, this signal separates the two regimes with mean similarity of 0.37 and 0.03 (Fig~\ref{fig:mitigation}). This gate effectively lifts the macro F1 on answered cases from 61.0 to 74.2, with 99\% of the answered cases being ones with correct evidence, suggesting that the gating mechanism on retrieval confidence serves as a valid proxy to downstream prediction performance. \textbf{Finding~3:} As the failure is tied to quality of evidence, we can teach a detector to abstain when it is unsure of the retrieval.

\subsection{Expert human ceiling}
\label{sub:human}
Three annotators with biomedical training independently labeled a stratified 120-pair sample, each
shown the question, the retrieved evidence snippet, and both answers in randomized order. Their
majority-vote macro F1 is \textbf{53.3} (per-annotator 55.0/54.2/57.5), barely above chance with
Fleiss' kappa of 0.42. The near-chance agreement suggests \emph{intrinsic difficulty}, whereas the fabrication is not easily identifiable by expert when the disconfirming evidence is absent from retrieved snippet. We include further details regarding expert annotation in Appendix~\ref{app:human}. 

\subsection{Consistent difficulty over judges}
\label{sub:committee}

While \ourdataset{}'s fabrications were selected by a single LLM as the pairwise difficulty filter, this raises concern that the benchmark is hard only for that judge. We therefore test with an order-invariant committee. For each pair we ask a judge which of the two answers is the fabrication under both answer orderings and count the pair solved only if the fabrication is identified regardless of position, a de-biasing that is essential because LLM judges are position-influened about half the time on the task with order consistency of about 51 to 54\%. We then employ two judges of different scale, i.e. Qwen3-4B and Qwen3-30B, which solve only \textbf{37.6\%} and \textbf{41.2\%} of pairs order-invariantly where theoretical randomness is 25\%, and their difficulty is \emph{correlated} with inter-judge Cohen's
kappa of 0.54. Appendix~\ref{app:committee} reports the judge position-bias analysis that motivates the order-invariant protocol. \textbf{Finding~4:} The difficulty is a property of the fabrications themselves and is consistent across judge used to select them.

\subsection{Case studies}
\label{sub:cases}
Table~\ref{tab:cases} shows four real \ourdataset{} pairs, one per fabrication type, of which
the negation flip is the running example from the introduction. Each is a single-token edit that
changes the clinical claim while leaving the sentence fluent, on-topic, and stylistically identical
to its ground truth, so none of them is falsifiable by reading alone. Together they illustrate the
detectability ordering of Section~\ref{sub:taxonomy}. The numeric and directional cases leave a token a
detector can check almost mechanically, as the passage either states a 60 HU threshold or it does
not, and it either reports a higher or a lower transfusion incidence. The entity swap is harder,
because \emph{IL-6} is as plausible a cytokine as \emph{TNF-$\alpha$} in this context, so the
detector has to locate which one the evidence actually names. The negation flip is hardest of all,
since \emph{with} and \emph{without} are equally fluent, the surrounding claim is untouched, and only
the polarity of the source statement settles it. In every case the decision rests on the passage
rather than on the answer, resonating with the relationship between detection and evidence quality in Finding~1 (Section~\ref{sub:evidence}), e.g. under a
mis-retrieved passage, about a different cohort, a different cytokine, or a different imaging
threshold, each of these edits passes unchallenged.

\begin{table}[t]
\centering
\small
\begin{tabular}{@{}p{0.18\columnwidth}p{0.72\columnwidth}@{}}
\toprule
Type & Question, and word-level edit (\textbf{ground truth} / \emph{fabrication}) \\
\midrule
Negation\newline(running ex.) & \emph{Do extremes of shock index predict death in trauma patients?}\newline ``\ldots increased mortality in individuals \textbf{with} / \emph{without} head injuries.'' \\
\addlinespace[2pt]
Entity & \emph{Does bariatric-surgery weight loss restore adipose PNPLA3 expression?}\newline ``\ldots otherwise inhibited by \textbf{TNF-$\alpha$} / \emph{IL-6}.'' \\
\addlinespace[2pt]
Numeric & \emph{Are low-attenuation plaque and the napkin-ring sign the strongest predictors of MACE?}\newline ``\ldots LAP values below \textbf{60} / \emph{50} HU and the presence of the NR sign are the strongest predictors\ldots'' \\
\addlinespace[2pt]
Directional & \emph{Is prophylactic plasma transfusion before interventional radiology associated with reduced bleeding?}\newline ``\ldots was linked to a \textbf{higher} / \emph{lower} incidence of periprocedural RBC transfusions.'' \\
\bottomrule
\end{tabular}
\caption{Four \ourdataset{} case studies, one per fabrication type, ordered hardest to easiest to detect (Section~\ref{sub:taxonomy}). A single-word edit changes the medical claim while preserving style, so each is identifiable only against the correct evidence.}
\label{tab:cases}
\end{table}

\section{Discussion and Research Opportunities}
\label{sec:discussion}
Our results reframe what progress on medical fabrication detection means: because detection is
governed by evidence \emph{correctness} rather than fabrication subtlety, the strong numbers reported
under gold evidence mask a below-chance collapse under \emph{wrong} evidence, so trustworthy
detection must be both \emph{evaluated} and \emph{deployed} under realistic, imperfect evidence. This
reframing opens four directions, developed in Appendix~\ref{app:opportunities}. \emph{Retrieval-aware,
evidence-robust detection} (Findings~1 and~3) is the central open problem: a detector that reasons about
evidence provenance and abstains when the passage is likely wrong, trained on \ourdataset{}'s paired
gold/none/distractor conditions rather than merely measured by them. \emph{Beyond snippet grounding}
(Section~\ref{sub:human}) follows from the expert ceiling, motivating full-document and multi-passage verification
in place of single-snippet entailment. \emph{Human-AI collaboration} (Finding~3) turns the
confidence gate into a triage whose cost and benefit under real clinician time is worth quantifying.
\emph{Co-evolutionary difficulty curricula} (Finding~4) would pit a fabrication generator against diverse detectors with difficulty scored on held-out judges, decoupling intrinsic
difficulty from difficulty against any one model. Extending \ourdataset{} to clinical notes,
other specialties, and non-English corpora remains important for external validity.

\section{Conclusion}
We introduced \ourdataset{}, a benchmark of word-level medical fabrications whose paired ground truths and fabrications share LLM authorship and near-identical surface form (median ROUGE-L 0.95), removing the authorship-style shortcut that inflates prior evaluations. Across supervised encoders, LLM judges, an NLI verifier, medical-tuned and reasoning models, and expert clinicians, the task is unsolved without gold evidence, where no detector clears about 60\% macro F1. Our central finding is that detection tracks evidence \emph{correctness}, not fabrication subtlety: under wrong evidence, detection falls below chance, so gold-evidence benchmarks overstate deployable performance. A one-line retrieval-confidence gate converts this failure into a safe abstention (61\% to 74\% macro F1). We release \ourdataset{}, all code, and every baseline to move medical fabrication evaluation toward the realistic, imperfect-evidence regime it must ultimately serve.

\section*{Limitations}
\label{sec:limitations}
\ourdataset{} carries five limitations, each expanded in Appendix~\ref{app:limitations}. Its
fabrications are LLM-generated over PubMedQA-derived answers rather than harvested from real clinical
text, so they do not capture the vocabulary, length, or information density of clinical notes. The
expert ceiling rests on a stratified 120-pair sample labelled by three biomedically-trained
annotators, which a larger panel with full-document evidence and licensed clinicians would tighten.
Our test that difficulty is intrinsic uses an order-invariant committee of two Qwen judges,
establishing stability across model \emph{scale} but not across model \emph{family}. The
evidence-degradation sweep probes the wrong-evidence regime adversarially because natural retrieval
on \ourdataset{}'s corpus is near-perfect, so we approximate rather than fully instantiate the
deployment curve. Finally, the retrieval-confidence gate uses a simple TF-IDF question-evidence
similarity, a deliberately minimal and model-free version that a learned retrieval-quality estimator
would likely improve on.

\section*{Ethical Considerations}
\label{sec:ethics}
\noindent\textbf{Releasing fabricated medical content.} \ourdataset{} contains 646 fluent,
clinically plausible statements that are factually wrong, which is exactly what makes it useful for
studying detection and also what makes it dual-use. Three properties of the release keep it oriented
toward detection. Every fabrication ships paired with its ground truth, its supporting evidence
passage, and its label, so no item can be mistaken for a verified claim. Every fabrication is a
single-word edit of an answer derived from published PubMedQA abstracts rather than a novel medical
assertion, so the release adds no false content beyond the altered token. The benchmark is
distributed for research on detecting errors that deployed medical LLMs already produce. The
fabricated statements are the objectionable content in this artifact, and their factual status is
documented rather than assumed: biomedically-trained annotators reviewed a stratified sample of 120
pairs and confirmed 85.8\% as carrying a genuine factual error, with the residual documented in
Section~\ref{sub:human} as a construct boundary rather than smoothed over. We judge the benefit to
detection research to outweigh the marginal risk, but we state the tension rather than treat the
release as neutral.

\noindent\textbf{Our results are a caution, not a capability.} No detector we evaluate clears about
60 macro F1 without a gold passage, expert clinicians given the retrieved snippet sit near chance,
and under a wrong passage detection falls below its own no-evidence baseline. Nothing reported here
is suitable for deployment as a clinical safety filter, and the retrieval-confidence gate of
Section~\ref{sub:mitigation} is a way to make evaluation realistic rather than a certified safeguard.
The paper should be read as evidence that automated medical fabrication detection is not yet
dependable, and that a system reporting high accuracy under gold evidence may fail badly once its
retrieval is wrong.

\noindent\textbf{Human annotation.} The expert study in Section~\ref{sub:human} involved three
annotators with biomedical training, each reviewing the same stratified sample of 120 pairs blind to
the automatic labels, with the two candidate answers shown in randomized order.
The instructions put to annotators are reproduced
verbatim in Appendix~\ref{app:human}, together with how they were recruited, whether and how they
were compensated, how consent was obtained, the ethics-board status of the protocol, and the
demographic and geographic characteristics of the annotator population. The task carries no risk to
annotators, since it involves only published biomedical abstract text. No annotator-identifying
information is released, and the per-annotator records in the data archive are keyed only as
\texttt{annotator\_a}, \texttt{annotator\_b}, and \texttt{annotator\_c}.

\noindent\textbf{Data provenance and privacy.} \ourdataset{} is derived from
MedHallu~\cite{pandit2025medhallucomprehensivebenchmarkdetecting}, itself built on
PubMedQA~\cite{jin-etal-2019-pubmedqa}, whose source material is published biomedical abstracts. The
corpus therefore contains no patient records, no clinical notes, and no personally identifying
information, and our pipeline introduces none, since it only rewrites and perturbs existing answer
text. Redistribution honours the upstream MedHallu and PubMedQA terms, which the released data
archive records alongside our own.

\noindent\textbf{Compute.} The work trains no large model. Dataset construction and the GPT-4o
baselines are API inference, the open models are served locally for inference, and the only fitted
models are a DeBERTa fine-tune and lightweight probes. Energy cost is dominated by inference over
1,292 answers across the evidence conditions and random seeds, which is small relative to a
pretraining or large-scale fine-tuning study.

\bibliography{reference}

\appendix
\section{Extended Related Work}
\label{app:relatedwork}
This section expands the condensed treatment in Section~\ref{sec:related}.

\paragraph{Hallucination datasets.} Early work by \citet{azaria-mitchell-2023-internal} marked the complete data-model pipeline by introducing a hidden-state-based classifier together with a simple true/false dataset, though it was limited to short synthetic statements. Subsequent datasets such as HaluEval~\cite{li-etal-2023-halueval} extended coverage to conversational hallucinations and inspired multi-type categorization frameworks~\cite{zhang2024knowhaluhallucinationdetectionmultiform,10.1145/3703155}. Building on this taxonomy, MedHallu~\cite{pandit2025medhallucomprehensivebenchmarkdetecting} provides a medical-domain benchmark classifying hallucinations into four types: question misinterpretation, incomplete information, mechanistic misattribution, and methodological fabrication. Fabrication cases nonetheless remain scarce, despite their critical importance in high-stakes domains such as healthcare, law, and journalism~\cite{li2023chatdoctor}, which motivates our construction of a fabrication-focused benchmark.

\paragraph{Detector families.} \emph{Uncertainty-based} methods quantify confidence through token probabilities~\cite{kuhn2023semantic}, verbalized self-assessments~\cite{lin-etal-2022-truthfulqa}, or response consistency across multiple generations~\cite{manakul-etal-2023-selfcheckgpt}, but they rely on latent spaces that do not inherently separate truthful from hallucinated responses. \emph{Supervised} approaches train classifiers on labeled hallucination data under the assumption that truthfulness is implicitly encoded within LLM internal states~\cite{azaria-mitchell-2023-internal}, but their scalability is constrained by costly human annotation. \emph{LLM-agentic} frameworks employ one or more LLMs as evaluators to verify factuality~\cite{huo2023retrievingsupportingevidencellms}, though they remain prone to self-hallucination and exhibit inconsistent performance~\cite{chiang2024chatbot}.

\paragraph{Retrieval-augmented verification.} A complementary line of work grounds factuality judgments in external evidence. \citet{shuster-etal-2021-retrieval-augmentation} show that augmenting LLM generation with retrieved passages substantially reduces hallucination rates in open-domain dialogue. In the biomedical setting, retrieval over PubMed-indexed corpora~\cite{jin-etal-2019-pubmedqa} provides a trusted evidence base for claim verification. Graph-based approaches such as GCA~\cite{10.1609/aaai.v39i22.34559} further model entity-relation structure to assess consistency between a claim and a knowledge graph. These methods treat retrieval as an input-augmentation step rather than an internal grounding constraint: the LLM is still free to generate tokens that deviate from the retrieved text. \ourmodel{} differs by prompting the LLM to copy words or phrases from the retrieved evidence at each masked slot (a measured 63.5\% verbatim token-copy rate, Appendix~\ref{app:weft}), anchoring reconstruction to verifiable source text and making deviation measurable rather than latent.

\paragraph{Human-in-the-loop validation.} Medical NLP systems benefit from domain-expert oversight to ensure annotations capture genuine clinical errors~\cite{singhal2025toward}. Our dataset construction incorporates a human-review stage in which medical annotators verify that generated fabrications contain genuine factual errors while preserving the stylistic form of the ground truth, preventing trivially detectable artifacts. This step is distinct from the automated ROUGE-L and SPPO filters: it provides a qualitative check that automated metrics cannot fully replicate.

\section{Benchmark Construction Details}
\label{app:construction}
This section expands the condensed construction summary in Section~\ref{sub:ourdataset}.

\paragraph{Curation and task formulation.} \ourdataset{} is constructed from MedHallu~\cite{pandit2025medhallucomprehensivebenchmarkdetecting}, a multi-domain medical QA benchmark spanning pharmacology, clinical oncology, rheumatology, and general internal medicine. All three pipeline stages (ground-truth rewriting, fabrication generation, and quality filtering) are executed with GPT-4o~\cite{openai2024gpt4ocard}. Fabrication detection is formulated as a binary classification task: given a question, supporting evidence, and a candidate answer, predict whether the answer is a ground truth (label~0) or a fabrication (label~1). To ensure dataset quality, a subset of generated fabrications was additionally reviewed by medical annotators to confirm factual inaccuracy without introducing vocabulary shifts that surface-level detectors could exploit.

\paragraph{The authorship-style bottleneck.} During dataset construction, we found that existing generation pipelines substantially alter the linguistic form of ground truths (Table~\ref{tab:medhallu-gptzero}), creating stylistic artifacts that models exploit as shortcuts instead of reasoning about factuality. Consequently, detectors achieve inflated accuracy by distinguishing writing styles rather than detecting genuine factual inconsistencies. To address these limitations, we propose a data-centric generation pipeline for the production of realistic, word-level fabrications that reflect genuine LLM failure modes. Specifically, we 1) \textbf{replace human-authored ground truths} with LLM-rewritten counterparts to align stylistic distributions, 2) generate \textbf{fabrications conditioned} on the rewritten ground truths to ensure factual grounding, and 3) \textbf{constrain the size of word-level alterations} to preserve sentence structure and distributional fidelity. The resulting dataset, \ourdataset{}, provides challenging but realistic fabrications with a median ROUGE-L of 0.95 between paired samples (Table~\ref{tab:dataset-stats}), capable of inducing misclassification in state-of-the-art models~\cite{katz2023gpt4}. 

\begin{table}[t] 
  \centering
  \setlength{\tabcolsep}{4pt}
  \begin{tabularx}{1\linewidth}{X}
  \toprule[0.5mm]
  \textbf{LLM-Generated Hallucination} \\
  \hline
  Targeting mid- and late-stage osteoblast differentiation markers (including ALP, osteocalcin, osteopontin, and bone sialoprotein) via RNAi may yield anticipated RNAi-nanogel nanostructured polymer-based prophylaxis for HO. \\
  \hline
  \textit{GPTZero Analysis:} 72\% AI, 0\% Mixed, 28\% Human \\
  \midrule[0.5mm]
  \textbf{Human-Written Ground Truth} \\
  \hline
  RNAi targeting mid- and late-stage osteoblast differentiation markers such as ALP, osteocalcin, osteopontin, and bone sialoprotein may produce the desired RNAi-nanogel nanostructured polymer HO prophylaxis.\\
  \hline
  \textit{GPTZero Analysis:} 3\% AI, 0\% Mixed, 97\% Human\\
  \bottomrule[0.5mm]
  \end{tabularx}
  \caption{An online LLM writing detector can distinguish between the ground truth and hallucination in MedHallu~\cite{pandit2025medhallucomprehensivebenchmarkdetecting}.}
  \label{tab:medhallu-gptzero}
  
\end{table}

\paragraph{Rewriting.} Figure~\ref{fig:generation-overview} shows an example from MedHallu~\cite{pandit2025medhallucomprehensivebenchmarkdetecting} and its LLM-rewritten counterpart. We prompt the LLM to rewrite the human-written ground truth without changing the semantic meaning of the context, seeking to transform the text's stylistic pattern to an LLM-authored one. The formal definition is given in Appendix~\ref{def:rewrite}.

\paragraph{Fabrication.} We generate word-level fabrications with the LLM conditioning on both the supporting evidence and the rewritten ground truth. The prompt includes relevant knowledge chunks and the correct answer to ensure factual grounding independent of the LLM's pre-trained knowledge. The formal definition is given in Appendix~\ref{def:fabricate}.

\paragraph{Quality control.} To mitigate distributional shifts between fabricated and ground-truth samples, we adopt a two-stage quality-control process (Algorithm~\ref{alg:fabrication_qc}) during fabrication that jointly evaluates semantic and structural similarity. Structural similarity is quantified using the ROUGE-L recall score~\cite{rouge}, where generations scoring below 0.7~\cite{Faray_de_Paiva2025.02.06.25321749} are discarded as structurally divergent. We then apply a preference-based difficulty filter inspired by self-play preference optimization~\cite{wu2024selfplaypreferenceoptimizationlanguage}: an LLM judge is presented with both the fabricated sample and the rewritten ground truth for the same question and asked which answer is factually correct. We retain only fabrications that the judge incorrectly prefers over the ground truth, since these represent genuinely deceptive cases that an LLM cannot easily detect. This process iterates until all samples satisfy both semantic and structural similarity constraints.

\begin{algorithm}[t]
\caption{Two-Step Fabrication Quality Control}
\label{alg:fabrication_qc}
\begin{algorithmic}[1]
\small
\REQUIRE Ground truth $\hat{x}^{gts}_{LLM}$, LLM, threshold $\tau_{str}=0.7$
\ENSURE Validated fabrication $\hat{x}^{fab}_{LLM}$
\REPEAT
    \STATE Generate $x^{fab}_{LLM}$ from $\hat{x}^{gts}_{LLM}$ using LLM
    \STATE Compute $r_{\hat{x}}=\text{ROUGE-L}(\hat{x}^{gts}_{LLM},x^{fab}_{LLM})$
    \IF{$r_{\hat{x}}<\tau_{str}$}
        \STATE Discard as structurally divergent
    \ELSIF{LLM judge prefers $\hat{x}^{gts}_{LLM}$ over fabrication}
        \STATE Discard as non-deceptive (easily detected)
    \ENDIF
\UNTIL{$x^{fab}_{LLM}$ passes structural \& semantic checks}
\RETURN $\hat{x}^{fab}_{LLM}=x^{fab}_{LLM}$
\end{algorithmic}
\end{algorithm}

\section{Formal Definitions for Generating \ourdataset{}}
This section provides formal definitions for rewriting (Appendix~\ref{def:rewrite}) and fabrication (Appendix~\ref{def:fabricate}) from Section~\ref{sub:ourdataset}. We denote by $\mathcal{Q}$, $\mathcal{K}$ and $\mathcal{X}$ the space of medical questions, retrieved knowledge sources, and natural-language answers respectively. 

\subsection{Rewriting} \label{def:rewrite} Given a question $q \in \mathcal{Q}$, knowledge $k \in \mathcal{K}$, and a human-authored answer $x^{\mathrm{gts}}_{\text{human}} \in \mathcal{X}$, the LLM rewriting operator $f_{\mathrm{LLM}} : \mathcal{Q} \times \mathcal{K} \times \mathcal{X} \to \mathcal{X}$ generates a rewritten ground truth $\hat{x}^{\mathrm{gts}}_{\mathrm{LLM}} = f_{\mathrm{LLM}}(q, k, x^{\mathrm{gts}}_{\text{human}})$, whose semantic content satisfies $\text{Content}(\hat{x}^{\mathrm{gts}}_{\mathrm{LLM}}) \approx \text{Content}(x^{\mathrm{gts}}_{\text{human}})$ and remains entailed by the retrieved evidence ($k \models \hat{x}^{\mathrm{gts}}_{\mathrm{LLM}}$). 

\subsection{Fabrication} \label{def:fabricate} Conditioned on the rewritten ground truth $\hat{x}^{\mathrm{gts}}_{\mathrm{LLM}}$ and context $(q, k)$, the fabrication operator $g_{\mathrm{LLM}} : \mathcal{Q} \times \mathcal{K} \times \mathcal{X} \to \mathcal{X}$ produces a fabricated answer $\hat{x}^{\mathrm{fab}}_{\mathrm{LLM}} = g_{\mathrm{LLM}}(q, k, \hat{x}^{\mathrm{gts}}_{\mathrm{LLM}})$, which diverges from factual correctness while preserving stylistic and structural similarity. This yields hard-negative variations which align with the sample’s linguistic distribution. 

\section{Supporting Experiments} 
\subsection{Fabrication has different data distribution from non-fabrication} 
\begin{SCtable}[][t] 
\centering
\small 
\resizebox{0.58\linewidth}{!}{
\begin{tabular}{lccccc}
\toprule[1pt]
\textbf{Metric} & \textbf{Split Type} & \textbf{Average} & \textbf{BERT} & \textbf{DeBERTa} & \textbf{ELECTRA} \\
\midrule
\multirow{2}{*}{\textbf{F1}}
& Random Split & 0.8433 & 0.7147 & 0.9411 & 0.8742 \\
& Fabrication & 0.5266 & 0.3333 & 0.9132 & 0.3333 \\
\multirow{2}{*}{\textbf{Precision}}
& Random Split & 0.7984 & 0.5781 & 0.9428 & 0.8742 \\
& Fabrication & 0.4712 & 0.2500 & 0.9137 & 0.2500 \\
\multirow{2}{*}{\textbf{Recall}}
& Random Split & 0.8376 & 0.7031 & 0.9421 & 0.8675 \\
& Fabrication & 0.6378 & 0.5000 & 0.9133 & 0.5000 \\
\bottomrule[1pt]
\end{tabular}
}
\caption{Performance degradation when classifiers trained on non-fabrication data are evaluated on fabrication-based datasets.} 
\label{tab:fab-vs-non-fab}
\end{SCtable} 
\subsection{Embedding model fails in highly structural similar sentences} 
\sidecaptionvpos{figure}{t} 
\begin{SCfigure}[][t]
\centering
\includegraphics[width=0.58\linewidth]{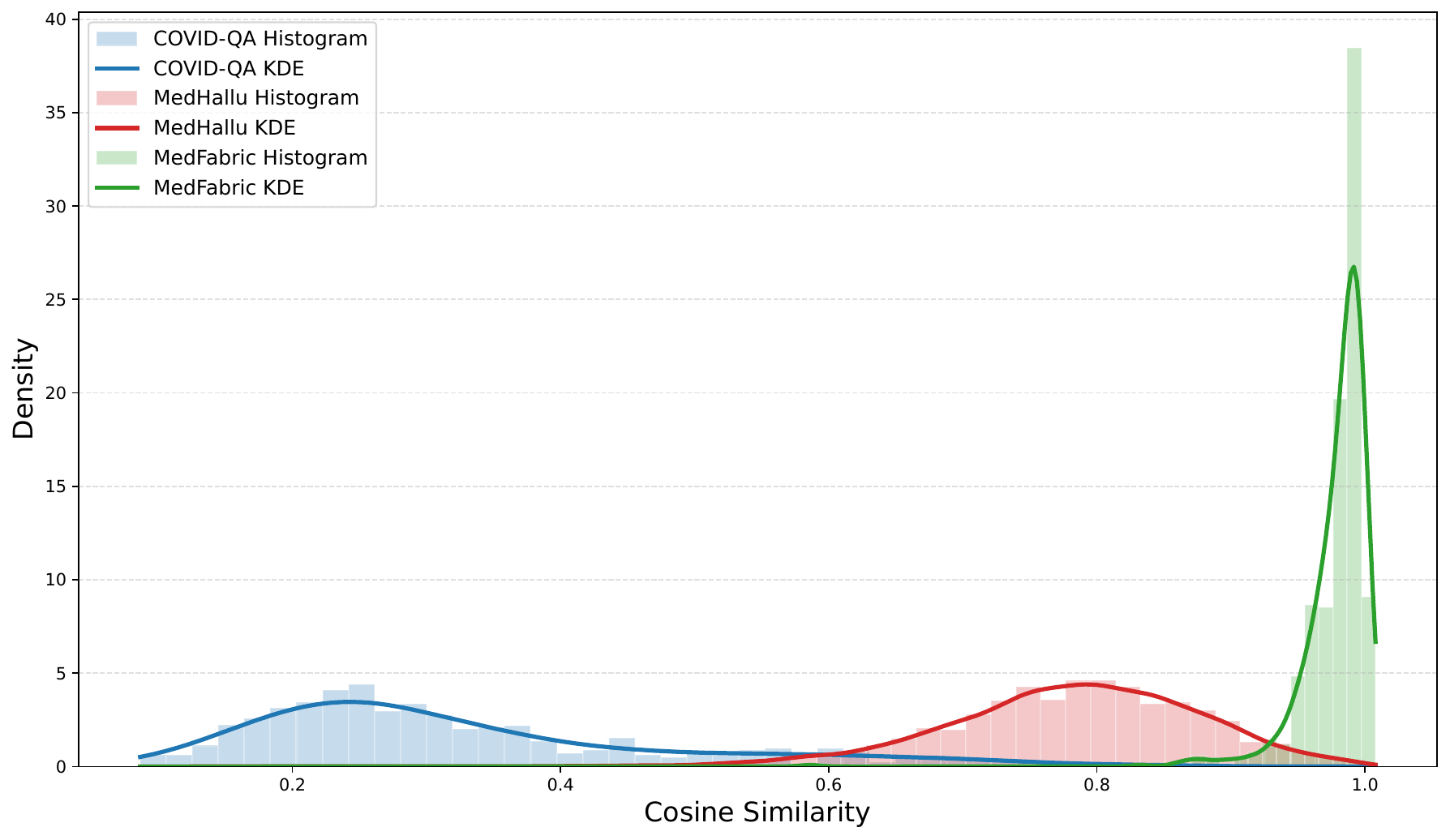}
\caption{Cosine similarity between ground truth and fabricated samples in \ourdataset{} is high compared to MedHallu~\cite{pandit2025medhallucomprehensivebenchmarkdetecting}, suggesting embedding-based model failures in distinguishing structurally similar sentences.}
\label{fig:similar-embedding}
\end{SCfigure}

\section{Full Evidence-Degradation Results}
\label{app:evsweep}
Table~\ref{tab:evsweep-full} expands Figure~\ref{fig:evsweep} with per-condition macro F1 (mean\,$\pm$\,std over three seeds) for both model scales on both benchmarks. Two regularities hold in every row: (i) correct evidence (gold or successfully retrieved) carries the task, and (ii) a wrong (distractor) passage drives detection \emph{below} the answer-only baseline, below chance (about 35) for the small model on both benchmarks. On \ourdataset{}, TF-IDF retrieval is near-perfect (recall@1 93 to 97\%), so the retrieved column nearly matches gold. This is an artifact of pairing each question with a single easily-matched passage. Enlarging the retrieval corpus 2.5-fold with 997 additional medical passages moves recall@1 only to 87\% (Qwen3-4B) / 91\% (GPT-4o) and detection by 1 to 2 points, confirming that on this benchmark \emph{natural} retrieval almost never fails and the realistic risk, wrong evidence, must be probed adversarially.

\begin{table}[t]
\centering
\small
\resizebox{\columnwidth}{!}{%
\begin{tabular}{llcccc}
\toprule
Bench. & Model & Gold & Retr.\ & None & Distr. \\
\midrule
\multirow{2}{*}{\ourdataset{}} & Qwen3-4B & 71.1\tiny{$\pm$5.8} & 68.2\tiny{$\pm$7.6} & 57.4\tiny{$\pm$1.9} & 35.0\tiny{$\pm$1.3} \\
 & GPT-4o & 91.3\tiny{$\pm$1.6} & 89.5\tiny{$\pm$2.5} & 60.2\tiny{$\pm$1.5} & 47.2\tiny{$\pm$3.1} \\
\midrule
\multirow{2}{*}{MedHallu} & Qwen3-4B & 78.6\tiny{$\pm$1.2} & n/a & 59.2\tiny{$\pm$5.2} & 34.7\tiny{$\pm$0.4} \\
 & GPT-4o & 74.7\tiny{$\pm$6.1} & n/a & 64.3\tiny{$\pm$7.3} & 50.0\tiny{$\pm$9.5} \\
\bottomrule
\end{tabular}}
\caption{Macro F1 by evidence condition (5-shot, 3 seeds). \emph{Retr.}\ = TF-IDF top-1 (recall@1 93\% Qwen3-4B / 97\% GPT-4o on \ourdataset{}). The distractor column is below the answer-only (\emph{None}) column in every row.}
\label{tab:evsweep-full}
\end{table}

The evidence-correctness effect is not confined to the two strongest detectors. Table~\ref{tab:gold-vs-none} contrasts gold-evidence and answer-only macro F1 across four model families and types, a frontier model, a small open model, a reasoning model, and a medical fine-tuned model, and every one gains substantially from correct evidence while none of them clears about 60 without it. The task difficulty lives in the evidence, not the model.

\begin{table}[t]
\centering
\small
\resizebox{\columnwidth}{!}{%
\begin{tabular}{lccc}
\toprule
Model & Gold & Answer-only & $\Delta$ \\
\midrule
GPT-4o (frontier) & 91.3 & 60.2 & +31.1 \\
Qwen3-4B (small) & 71.1 & 57.4 & +13.7 \\
R1-Distill-7B (reasoning) & 62.4 & 51.6 & +10.8 \\
BioMistral-7B (medical) & 40.2 & 34.3 & +5.9 \\
\bottomrule
\end{tabular}}
\caption{Gold-evidence vs.\ answer-only macro F1 on \ourdataset{}, 5-shot. All four model types benefit from correct evidence. None is deployable without it.}
\label{tab:gold-vs-none}
\end{table}

\section{Bootstrap Significance}
\label{app:bootstrap}
The leaderboard margins in Table~\ref{tab:leaderboard} are validated by a paired bootstrap (10,000 resamples over the 646 test pairs, resampling pairs with replacement and recomputing macro F1 for both detectors on each resample). \ourmodel{} attains 56.9 macro F1 (95\% CI [54.3, 59.3]) and the graph-based GCA detector 47.2 (95\% CI [44.6, 49.8]). The paired difference is 9.7 (95\% CI [6.1, 13.3], $p<0.001$). \ourmodel{} thus significantly leads prior zero-shot detectors while still failing to clear about 60, the two claims that anchor the paper.

\section{Committee Difficulty and LLM-Judge Position Bias}
\label{app:committee}
\ourdataset{} was filtered with a single LLM (GPT-4o) as the pairwise difficulty judge, raising the concern that the benchmark is hard only for that judge. We re-audit difficulty with an \emph{order-invariant} committee of two Qwen judges spanning a 7.5-fold scale gap, counting a pair solved only if the fabrication is identified under \emph{both} answer orderings (Table~\ref{tab:committee}). Both judges solve well under half of the pairs. Their difficulty is correlated (kappa of 0.54, where a pair hard for the small judge is hard for the large one 79\% of the time vs.\ a 59\% base rate, and 50\% of all pairs are hard for both). Scaling the judge does not erase the difficulty and the hard pairs are largely the same pairs.

\begin{table}[t]
\centering
\small
\begin{tabular}{lcc}
\toprule
Judge & Order-invariant solve\,\% & Params \\
\midrule
Chance floor & 25.0 & n/a \\
Qwen3-4B & 37.6 & 4B \\
Qwen3-30B-A3B & 41.2 & 30B \\
\bottomrule
\end{tabular}
\caption{Order-invariant committee solve rate on \ourdataset{} (inter-judge kappa of 0.54). Neither capable judge clears about 41\%.}
\label{tab:committee}
\end{table}

The order-invariance requirement is not cosmetic. LLM pairwise judges on this task are strongly \emph{position}-biased: a Phi-family judge picks the first-presented answer 97\% of the time, and a Yi-1.5-9B judge agrees with itself across the two orderings only 24\% of the time (10.2\% order-invariant-correct), both are unusable as difficulty judges. Even the Qwen judges are order-consistent only 51 to 54\% of the time, so a naive single-ordering protocol rewards positional guessing and would badly understate difficulty. This position bias is also why a cross-family held-out judge is left to future work: the available local cross-family models are too biased to serve as judges, and API budget for GPT-4o as a held-out judge was exhausted.

\section{Retrieval-Confidence Gate: Full Risk-Coverage}
\label{app:mitigation}
Table~\ref{tab:riskcov} gives the full risk-coverage curve behind Figure~\ref{fig:mitigation}. On a pool that mixes gold and distractor evidence in equal parts (a retriever right half the time), TF-IDF question-passage cosine similarity cleanly separates the two regimes (mean 0.37 gold vs.\ 0.03 distractor). The ungated detector, forced to answer every case, scores 61.0 macro F1, dragged down by the wrong-evidence half, which alone scores 33.4 (below chance). Abstaining on low-confidence evidence lifts macro F1 on the answered set monotonically toward the gold-evidence ceiling. At about 50\% coverage the answered set is 99\% correct-evidence cases and macro F1 reaches 74.2. A one-line confidence check thus converts the below-chance wrong-evidence failure into a safe abstention.

\begin{table}[t]
\centering
\small
\begin{tabular}{cc}
\toprule
Coverage\,\% (answered) & Macro F1 on answered \\
\midrule
100 (ungated) & 61.0 \\
76 & 65.7 \\
60 & 70.7 \\
54 & 72.8 \\
52 & 73.3 \\
50 & 74.2 \\
\bottomrule
\end{tabular}
\caption{Risk-coverage under mixed (gold/distractor) evidence, gating on TF-IDF retrieval confidence. Lower coverage gives higher macro F1 on the answered set as wrong-evidence cases are deferred to human review.}
\label{tab:riskcov}
\end{table}

\section{Human Validation Protocol}
\label{app:human}
\paragraph{Construct validity (Section~\ref{sub:ourdataset}).} Because \ourdataset{} targets medical factuality, automated ROUGE-L and SPPO filters alone cannot certify that a fabrication contains a \emph{genuine} factual error rather than a benign paraphrase. We therefore conducted an expert human-validation study. Three annotators with biomedical training independently reviewed a stratified sample of 120 fabrication and ground-truth pairs (18.6\% of \ourdataset{}), drawn evenly across ROUGE-L quartiles. Because most fabrications are minimal edits of their ground truth (median ROUGE-L 0.95), this sample spans the benchmark's full similarity range. Each annotator was shown the question, the supporting evidence, and the two candidate answers in randomized order, blind to the automatic labels, and judged for each answer whether it contained a genuine factual error rather than an acceptable rephrasing.

Raw inter-annotator agreement on the fabrication judgment, computed jointly across all three annotators, was Fleiss' kappa of 0.42 (Krippendorff's alpha of 0.43, moderate). The disagreements concentrated in mid-similarity rewrites where the contradicting evidence was not contained in the retrieved snippet shown at annotation time. We therefore adjudicated the 73 disputed pairs against the full source document. After adjudication, 85.8\% of sampled fabrications were confirmed to carry a genuine factual error while preserving the stylistic form of the ground truth, supporting the construct validity of the benchmark.

The residual 14\% are predominantly \emph{unsupported additions}, plausible claims the evidence neither states nor contradicts, which we count as hallucinations by construction but which human readers do not flag as outright errors, delineating the benchmark's construct boundary. Full annotation guidelines, the per-pair judgments, and the adjudication protocol are released with the dataset.

\paragraph{Detection ceiling (Section~\ref{sub:human}).} Three annotators with biomedical training independently labeled a stratified 120-pair sample of \ourdataset{} v2.2, each shown the question, the retrieved evidence snippet, and both answers in randomized order, and asked to mark which answer is fabricated. Stratification balanced medical specialty and fabrication type. Per-annotator macro F1 is 55.0 / 54.2 / 57.5 and the majority vote is 53.3, barely above chance, with Fleiss' kappa of 0.42 (Krippendorff's alpha of 0.43). The near-chance agreement reflects intrinsic difficulty rather than annotator noise: when the disconfirming evidence is absent from the retrieved snippet, the fabrication is not identifiable by expert reading, which bounds what any snippet-grounded detector can achieve.

To confirm the residual \emph{unsupported-addition} pairs (those adjudication did not certify as outright contradictions) do not drive the result, we restrict to the human-confirmed contradiction-only subset (106/120). This changes \ourmodel{} by only 0.7 points (55.6 to 56.3) and the human ceiling by 5.2 points (53.3 to 58.5). Both stay in the mid-50s, and the larger human gain confirms those pairs are disproportionately ambiguous to readers rather than mislabeled noise.

\paragraph{Instructions given to annotators.} Each annotator ran a local
single-page tool that presented the 120 sampled pairs one at a time. For every pair the interface
showed, in this order, the question, the supporting evidence passage under the heading ``Supporting
evidence (PubMed)'', then ``Answer A'', then ``Answer B'' under the note ``(highlighted = differs
from A)'' with the differing tokens highlighted inline. Which of the two answers held the ground
truth was randomised per pair and no automatic label was shown. The three questions put to the
annotator were, verbatim:

\begin{quote}\small
\textbf{Q1.} Judging \textbf{Answer A} against the evidence, is it: \emph{Factually correct} /
\emph{Contains a factual error} / \emph{Unsure}

\textbf{Q2.} Judging \textbf{Answer B} against the evidence, is it: \emph{Factually correct} /
\emph{Contains a factual error} / \emph{Unsure}

\textbf{Q3.} (optional) If A and B differ only by rephrasing with no factual change, mark here:
\emph{Differ only by paraphrase} / \emph{A real factual difference}
\end{quote}

\noindent A free-text ``Notes (optional)'' box accompanied each pair, progress was saved after every
pair, and sessions were resumable. No time limit was imposed and no risk to annotators arises from
the task, which involves only published biomedical abstract text. The tool is released as
\texttt{annotation\_tool/annotate.py} in the software archive.

\paragraph{Annotator recruitment and compensation.} None of the three annotators is an author of
this submission. They were recruited by cold email to individuals with biomedical training, and
participation was voluntary in that recipients were free not to reply. Annotators were compensated
at USD~50 per hour, a rate substantially above the minimum wage of the jurisdiction in which they
were recruited, so that participation was not financially coerced.

\paragraph{Consent and annotator population.} Before any annotation began, each annotator joined a
call in which the task, its purpose, and the intended use of the annotations in a research
publication were explained in detail, and consent to participate was obtained during that call.
Identifying information was collected only at recruitment, solely to confirm the annotator's
biomedical background, and was discarded immediately afterwards. No participant identity is
disclosed, and the released per-annotator records are keyed only as \texttt{annotator\_a},
\texttt{annotator\_b}, and \texttt{annotator\_c}. All three annotators are based in the United
States and have biomedical training.

\paragraph{Ethics review.} No ethics review was sought for this annotation study. It collects no
personal data beyond the recruitment check described above, which was discarded immediately, it
involves only published biomedical abstract text rather than patient or clinical material, and it
poses no risk to participants.

\section{Baseline Configurations and Reproducibility}
\label{app:baselines}
All detectors are evaluated on \ourdataset{} v2.2 (646 pairs / 1,292 answers) with a single consistent scoring rule (a forced \texttt{fabricated}/\texttt{gt} label scored by \texttt{classification\_report}). Few-shot exemplars are drawn only from the train split and evaluation is on the held-out test split. Few-shot LLM results report mean\,$\pm$\,std over three random seeds (varying the sampled exemplars and generation).
\begin{itemize}
\item \textbf{Few-shot LLMs.} GPT-4o (\texttt{gpt-4o}) via API. Qwen3-4B and Qwen3-30B-A3B are served locally with vLLM. 5-shot unless noted.
\item \textbf{Reasoning / medical baselines.} DeepSeek-R1-Distill-Qwen-7B (served at \texttt{max-model-len} 4096) and BioMistral-7B (system prompt merged into the user turn, as Mistral rejects a standalone system role).
\item \textbf{NLI.} DeBERTa-v3-large fine-tuned on MNLI, given the gold passage as premise and each answer as hypothesis. Contradiction maps to \texttt{fabricated} and entailment to \texttt{gt}.
\item \textbf{Prior detectors.} GCA (graph-consistency), TSV (task steering vector), and a supervised attention-probe (logistic regression on hidden states) reproduced on v2.2.
\item \textbf{\ourmodel{}.} Zero-shot Text2Table, entity check, dual word masking and filling, then a hybrid embedding-plus-LLM verdict, with Qwen3-Embedding-8B. Evidence is consumed only at the reconstruction step, never at the verdict.
\item \textbf{Significance.} Paired bootstrap, 10,000 resamples over the 646 test pairs (Appendix~\ref{app:bootstrap}).
\end{itemize}

\section{Analytical Experiments: What Makes \ourdataset{} Hard}
\label{app:analysis}
The following analyses dissect \emph{why} the benchmark is hard, using only the released pairs and the per-answer verdicts of a representative evidence-grounded LLM detector. They (i) quantify the authorship-style shortcut that \ourdataset{} removes, (ii) characterise the fabrication types, and (iii) rule out the two most obvious alternative explanations of difficulty.

\subsection{The authorship-style shortcut, quantified}
\label{app:shortcut}
We test whether the fabrication can be identified by \emph{surface style alone}, with no factual reasoning. For each benchmark we pool all answers, label each as ground-truth or fabrication, and train two classifiers under 5-fold stratified cross-validation: a \emph{style-only} model on stylometric features (character/word/sentence lengths, type-token ratio, punctuation, function-word and hedging ratios) and a \emph{lexical} model on TF-IDF 1- and 2-grams. Chance is 50\% balanced accuracy. Table~\ref{tab:shortcut} shows the shortcut is large on prior data and closed by \ourdataset{}'s LLM rewrite: on MedHallu a lexical classifier reaches \textbf{88.1\%} and a style-only classifier \textbf{80.4\%}, one can pick the fabrication without reading for facts. Pairing the \emph{human} ground truth against the LLM fabrication (as pre-rewrite versions of \ourdataset{} and most prior benchmarks do) is separable at 66 to 72\%. Once both answers share LLM authorship (released \ourdataset{}), every surface model falls to chance or below (31 to 56\%). The residual 55.9 of the linear style model is a length artifact (a minority of fabrications add a clause), which the tree and lexical models, free to exploit any real surface cue, do not convert into signal. Surface similarity is correspondingly high: median ROUGE-L 0.95 on \ourdataset{} versus 0.23 on MedHallu.

\begin{table}[t]
\centering
\small
\resizebox{\columnwidth}{!}{%
\begin{tabular}{lccc}
\toprule
Pairing & Style & Lexical & Med.\ ROUGE-L \\
\midrule
MedHallu (human GT vs LLM) & 80.4 & 88.1 & 0.23 \\
\ourdataset{} human GT vs LLM & 71.6 & 69.7 & 0.54 \\
\ourdataset{} LLM GT vs LLM (final) & 41.7 & 31.0 & 0.95 \\
\bottomrule
\end{tabular}}
\caption{Surface-only separability (balanced accuracy, chance is 50, with the best of style-only LR/GBM shown in \emph{Style}). A style/lexical classifier separates fabrication from GT at up to 88\% on prior data, but at or below chance once both answers share LLM authorship, the authorship-style shortcut is removed.}
\label{tab:shortcut}
\end{table}

\subsection{Fabrication taxonomy}
\label{app:taxonomy}
We categorise each fabrication by the ground-truth-to-fabrication edit (token diff): a changed number/dose, a flipped negation, a swapped direction/comparator, an added unsupported clause, or an entity/term substitution (Table~\ref{tab:taxonomy}). \ourdataset{} is dominated by entity/term swaps (54\%) and numeric/dosage changes (31\%). Per-type catch rates from the representative detector (545 matched fabrications, a relative pattern rather than a headline number) show the intuitive ordering: single-token \emph{negation flips} are the hardest to catch (58.8\%), while numeric and directional changes are the easiest (76 to 82\%).

\begin{table}[t]
\centering
\small
\begin{tabular}{lcc}
\toprule
Fabrication type & Share\,\% & Caught\,\% \\
\midrule
Entity/term swap & 54.2 & 66.2 \\
Numeric/dosage & 30.5 & 76.5 \\
Negation flip & 10.7 & 58.8 \\
Directional/comparator & 2.6 & 82.4 \\
Added unsupported clause & 2.0 & 66.7 \\
\bottomrule
\end{tabular}
\caption{Fabrication-type distribution in \ourdataset{} and per-type catch rate of a representative evidence-grounded detector. Negation flips are hardest and numeric/directional edits easiest.}
\label{tab:taxonomy}
\end{table}

\subsection{Detectability is not explained by subtlety or surface overlap}
\label{app:drivers}
If difficulty came from the fabrication being a \emph{subtler} edit, small edits would be caught less often. They are not. The number of changed tokens is uncorrelated with detection (point-biserial $r=+0.02$, $p=0.72$, catch rate 73/66/70/60\% across edits of 1 / 2 to 3 / 4 to 10 / more than 10 tokens). The fabrication's lexical overlap with the provided evidence is likewise uncorrelated ($r=+0.02$, $p=0.60$). Neither the subtlety of the edit nor its surface resemblance to the evidence predicts whether a detector catches it, ruling out the two most obvious alternative explanations and leaving evidence \emph{correctness} (Section~\ref{sub:evidence}) as the operative factor. Consistent with this, a learned steering-vector detector (TSV) attains AUC 0.47 on a 260-answer sample, its confidence scores are non-discriminative on \ourdataset{}.

\section{\ourmodel{} Detector: Full Pipeline}
\label{app:weft}
\begin{figure*}[t]
    \centering
    \includegraphics[width=\textwidth]{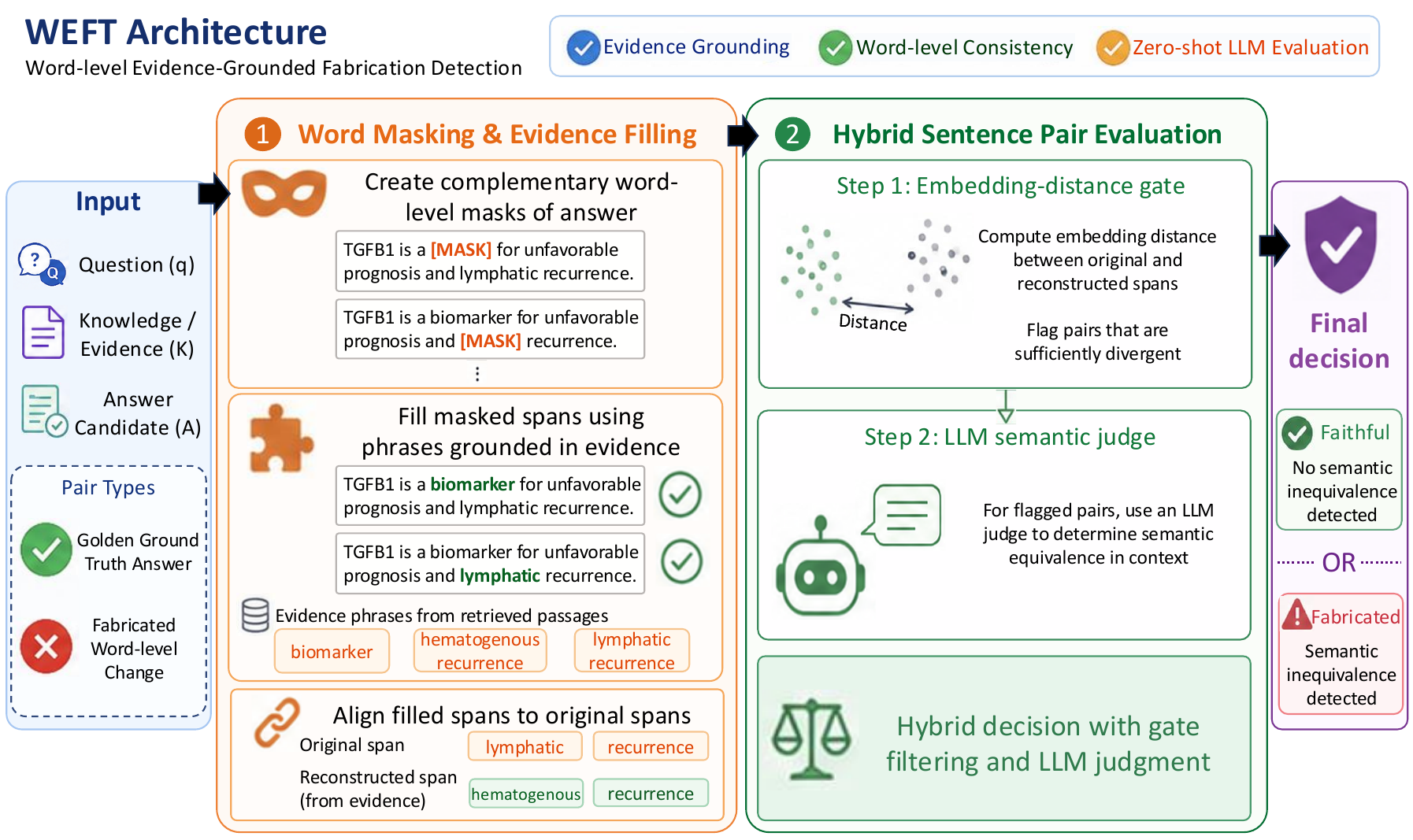}
    \caption{\ourmodel{} centers on two core modules: \textbf{(1) Word Masking and Filling}, which masks content words in the candidate answer and refills them from the retrieved evidence to expose word-level factual divergence, and \textbf{(2) Hybrid Sentence Pair Evaluation}, which verifies each reconstructed pair with an embedding-distance gate followed by an LLM semantic judge for low-variance decisions. An optional Text2Table Decomposition stage pre-structures longer multi-sentence inputs but is net-neutral on \ourdataset{}'s single-sentence answers.}
    \label{fig:detector}
\end{figure*}

\ourmodel{} is the zero-shot structured detector we report as a baseline (Section~\ref{sub:ourmodel}). It has two core modules and one optional pre-processing stage (Figure~\ref{fig:detector}).

\paragraph{Text2Table decomposition (optional).} This stage converts a medical statement into a structured table of biomedical entities and their contextual relations, linking each entity to its description (e.g., ``TGFBI expression'' maps to ``unfavorable lymphatic recurrence''). Complex sentences are decomposed into multiple entity-context rows and pronouns are resolved to explicit references. It benefits longer multi-sentence inputs but, as the leaderboard ablation shows (Section~\ref{sub:evidence}), is net-neutral on the single-sentence answers in \ourdataset{}, so we treat it as optional.

\paragraph{Word masking and filling.} From the entity-description table, two complementary masked tables are generated by alternately masking non-overlapping subsets of content words, so every content word is hidden exactly once across the two tables. Each masked slot is filled by an LLM conditioned on the \emph{gold} evidence passage bundled with each sample (one passage per sample, supplied by the source benchmark and reused verbatim at inference, so retrieval recall is not a confound in the main results. The deployment regime is studied through the knowledge ablation in Section~\ref{sub:evidence}). The LLM is prompted to copy exact words or phrases from the evidence rather than generate new content. Auditing compliance across 1,204 filled slots (40 pairs, GPT-4o fill), 63.5\% of filled content tokens appear verbatim in the evidence and 33.9\% of slots are fully grounded. Compliance is near-identical for ground-truth and fabrication fills (64.8\% vs.\ 62.2\% verbatim), so the copy rate itself carries no class signal: the discriminative information lies in \emph{where} the reconstruction diverges from the original.

\paragraph{Hybrid sentence-pair evaluation.} The two filled tables $T$ and $T'$ satisfy an invariant: for every slot $s$, exactly one of $T[s]$, $T'[s]$ holds the original answer value and the other holds the evidence-grounded reconstruction. Slot representations from Qwen3-Embedding-8B~\cite{qwen3embedding} are L2-normalised, so the slot-wise Euclidean distance between $T[s]$ and $T'[s]$ measures agreement between the answer and the evidence. Any non-zero distance flags the slot for LLM review (threshold of zero). Flagged slots are passed to an LLM that issues a single verdict. Embedding distance alone is near-random on \ourdataset{}: three off-the-shelf embedding models (BGE-M3~\cite{chen2024bge}, MiniLM-L6-v2~\cite{wang2020minilm}, and Qwen3-Embedding-0.6B~\cite{qwen3embedding}) reach only 41 to 48\% macro F1 as standalone classifiers, and near-identical ROUGE-L means almost no slot reconstructs to an exact match, so the gate forwards nearly all slots and the LLM verdict does the substantive work, with the gate's compute savings materialising only on lower-similarity inputs.

\section{Limitations in Detail}
\label{app:limitations}
This section expands the summary in Section~\ref{sec:limitations}.

\textbf{Synthetic benchmark.} \ourdataset{}'s fabrications are LLM-generated (evidence-conditioned single edits) over PubMedQA-derived answers rather than harvested from real clinical text. This gives control over the factual perturbation but does not capture the vocabulary, length, or information density of clinical notes, discharge summaries, or non-English corpora. MEDEC~\cite{medec2024}, which lacks per-instance evidence passages, is outside the evidence-grounded setting we study.

\textbf{Human study scale.} The expert ceiling (53.3\%) rests on a stratified 120-pair sample labelled by three biomedically-trained annotators (Fleiss' kappa of 0.42), each shown only the retrieved evidence snippet. A larger panel, full-document evidence, and licensed clinicians would tighten this estimate, though our contradiction-only re-scoring shows the near-chance result is robust to the construct-boundary pairs.

\textbf{Committee family diversity.} Our test that difficulty is intrinsic uses an order-invariant committee of two Qwen-scale judges (kappa of 0.54), establishing stability across model \emph{scale}. A capable cross-\emph{family} held-out judge (e.g., GPT-4o, the model the benchmark was filtered against) would make the claim airtight. Smaller non-Qwen models we tried (Phi-3.5, Yi-1.5) proved too position-biased to judge the pairwise task reliably.

\textbf{Retrieval realism.} The evidence-degradation sweep probes the \emph{wrong}-evidence regime adversarially (a mis-retrieved passage). On \ourdataset{}'s corpus, natural TF-IDF retrieval is near-perfect (recall@1 93 to 97\%) so the retrieved point nearly matches gold. Measuring the full deployment curve requires a large, noisy clinical evidence corpus, which we approximate but do not fully instantiate.

\textbf{Mitigation signal.} The retrieval-confidence gate uses a simple TF-IDF question-evidence similarity. A learned retrieval-quality estimator or calibrated abstention would likely trace a stronger risk-coverage frontier. We report the minimal, model-free version to show the effect is real and cheap.

\section{Research Opportunities in Detail}
\label{app:opportunities}
This section expands the directions sketched in Section~\ref{sec:discussion}.

Our results reframe what progress on medical fabrication detection means. Because detection quality is governed by evidence \emph{correctness} rather than fabrication subtlety, the strong numbers reported under gold evidence mask a below-chance collapse under \emph{wrong} evidence, so gold-evidence evaluation systematically overstates deployable performance. The practical consequence is that trustworthy detection must be both \emph{evaluated} and \emph{deployed} under realistic, imperfect evidence. Our retrieval-confidence gate is a first, model-free step toward the latter. This reframing opens several directions.

\noindent\textbf{Research opportunities.} Our findings point to concrete directions for the field.
\emph{(1)~Retrieval-aware, evidence-robust detection} (Findings~1 and~3): because detection tracks evidence \emph{correctness}, the open problem is a detector that estimates and reasons about evidence provenance, calibrating on retrieval quality and abstaining or down-weighting when the passage is likely wrong, extending our model-free confidence gate to a learned risk-coverage frontier. \ourdataset{}'s paired gold/none/distractor conditions further supply a natural training signal for fine-tuning detectors that resist the wrong-evidence collapse rather than merely measuring it, a direct route from benchmark to improved model.
\emph{(2)~Beyond snippet grounding} (Section~\ref{sub:human}): expert readers fail largely because the disconfirming fact is absent from the retrieved snippet, motivating full-document and multi-passage verification rather than single-snippet entailment.
\emph{(3)~Human-AI collaboration} (Finding~3): the confidence gate defines a natural triage, auto-verify confident cases, route the rest to clinicians, whose cost/benefit under real clinician time is worth quantifying.
\emph{(4)~Co-evolutionary difficulty curricula} (Finding~4): a curriculum in which a fabrication generator and a diverse committee of detectors are iteratively improved against one another, with fabrication genuineness anchored to retrieved evidence and human validation, and difficulty scored on held-out detectors never used during curation, would measure hardness against an unseen committee rather than a single judge, decoupling the \emph{intrinsic} difficulty of a fabrication from its difficulty against any one model and directly addressing the single-judge selection bias.
Finally, extending \ourdataset{} to real clinical notes, other specialties, and non-English corpora, and measuring downstream impact as a post-generation filter in clinical QA pipelines, remain important for external validity.

\end{document}